\pdfoutput=1

\documentclass[11pt]{article}

\usepackage[final]{acl}

\usepackage{times}
\usepackage{latexsym}
\usepackage{natbib}
\usepackage{float}
\usepackage{amssymb}
\usepackage{amsmath}
\usepackage{amsthm}
\usepackage{booktabs}
\usepackage{enumitem}
\usepackage{color}
\usepackage{multirow}
\usepackage{graphicx}
\usepackage[absolute,overlay]{textpos}

\usepackage[T1]{fontenc}

\usepackage[utf8]{inputenc}

\usepackage{microtype}

\usepackage{inconsolata}

\usepackage{graphicx}

\title{Cross-Modal Emotion Understanding: A Transformer-GAT Approach for Dialogue Emotion Recognition}

\author{
    Jiaqi Qiao, Yifan Lyu, and Xiujuan Xu\textsuperscript{*}
    \\
    Dalian University of Technology
    \\
    \texttt{qiaobright@mail.dlut.edu.cn};\quad
    \texttt{stevelyu811@gmail.com}
    \\
    \textsuperscript{*}\textbf{Correspondence:}
    \texttt{xjxu@dlut.edu.cn}
}
\begin{document}
\maketitle
\begin{abstract}
Multimodal emotion recognition is a key research area in affective computing, with applications in sentiment analysis, intelligent customer service, and human-computer interaction. However, existing methods often rely on single-modal features or simple multimodal fusion, failing to capture the synergy between global and local contexts, which limits model performance and emotion understanding. To address this challenge, we propose Transformer-GAT, a hybrid framework that combines Transformer and the Graph Attention Network to enable cross-modal emotion understanding. The Transformer is used to capture global semantic information, while the Graph Attention Network is employed to model fine-grained relationships between modalities, thereby enhancing the representation of emotional features. Experiments on the IEMOCAP and MELD datasets show that our model achieves weighted F1 scores of 72.45\% and 77.37\%, outperforming state-of-the-art methods. These results demonstrate that Transformer-GAT effectively integrates multimodal features, balances global and local contexts, and provides deeper emotional insights, offering new directions for multimodal emotion computing.
\end{abstract}

\section{Introduction}
Emotion recognition  as a key technology in human-computer interaction \citep{10.1007/978-3-540-30568-2_27,10.1007/978-3-642-11721-3_2}, enables more natural interaction between people and robots \citep{introduction1,introduction2}, addressing the challenge of providing human-like services to machines. Inspired by human empathy, machines should have the ability to detect emotions before offering services. Human communication involves multiple modalities, including text, video, and audio, which adds complexity to emotion recognition tasks. Therefore, multimodal emotion recognition (ERC) integrates information from text content, video, and audio signals to automatically identify the emotional state of speakers in conversations, and has been widely applied in areas such as chatbots \citep{introduction3} and sentiment mining on social media \citep{introduction4}.
\begin{figure}[h]
	\centering
	\includegraphics[width=0.48\textwidth]{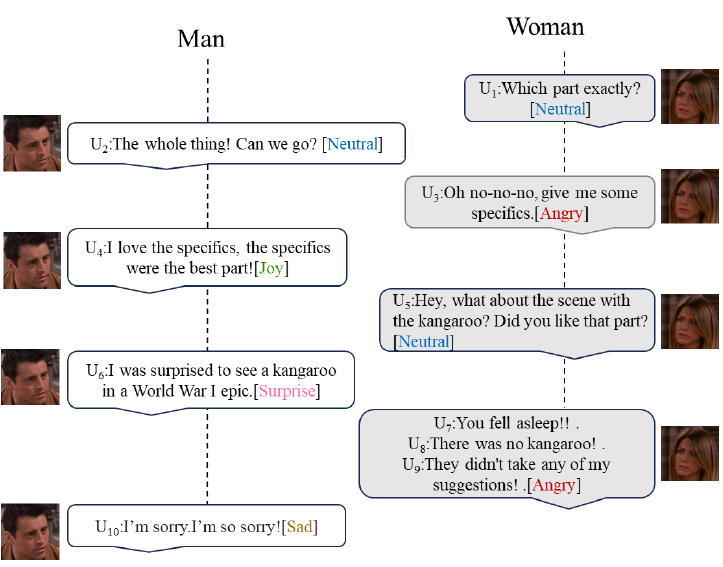}
	\caption{ A sample dialogue snippet from the MELD dataset.}  
    \label{fig:diaglue}
\end{figure}

Currently, emotion recognition primarily relies on single-modal features \citep{Bert-erc} or simplistic multimodal fusion methods \citep{COGMEN}. These approaches do not fully exploit the synergy between global semantics and local contexts, thereby limiting the depth of emotional understanding achieved by the models. Existing models often struggle to simultaneously capture long-distance, local, and short-distance contextual information when addressing emotion recognition tasks.

Although emotion recognition has made some progress in multimodal dialogue \citep{openvocabulary}, effectively integrating multimodal information and modeling the interaction between global and local contexts remains a critical unresolved challenge. In particular, existing research has yet to comprehensively address the modeling of long-distance and short-distance emotional associations.

To address these challenges, we propose a Transformer-GAT model that integrates the Transformer and Graph Attention Network \citep{velivckovic2017graph}. This model leverages the Transformer to capture global semantic information while employing GAT to construct long-distance, local, and short-distance contextual information, as illustrated in Figure \ref{fig:diaglue}, thus enhancing the representation of emotional features, please refer to Appendix \ref{app:image_explanation}. The effectiveness of the proposed model is validated through experiments on the IEMOCAP and MELD datasets.

The main contributions of this paper are as follows:
\begin{itemize}     
\item We propose a novel emotion recognition model, Transformer-GAT, which effectively integrates multimodal information and accurately captures long-distance, local, and short-distance contextual information. Additionally, our source code will be released on GitHub. 
\end{itemize}
\begin{itemize}      
\item The proposed model outperforms state-of-the-art methods in emotion recognition tasks, achieving significant improvements, particularly in weighted F1 scores.
\end{itemize}
\begin{itemize}       
\item This study offers new technical insights and research directions for the field of affective computing, laying a solid foundation for future exploration. 
\end{itemize}

The paper continues with a review of related work on emotion recognition in Section 2. Section 3 details our proposed model. Section 4 covers the experimental setup, followed by the results, analysis, and visualization in Section 5. The paper concludes in Section 6.


\section{Related Work}

ERC is crucial in dialogue systems, as it significantly enhances human-computer interactions and user experience. By accurately identifying emotions within dialogues, the system can provide more appropriate and empathetic responses, making interactions more natural and enjoyable. ERC involves modeling historical information to capture complex dependencies and time-series relationships within and between modalities, thereby extracting comprehensive contextual features. Current ERC research mainly includes methods based on Recurrent Neural Networks (RNNs)  \citep{ICON}, Transformers \citep{EmoBERTa}, Graph Neural Networks (GNNs)  \citep{DialogueCRN,LSDGNN,ijcai2026p647}, and novel approaches using large language models (LLMs)  \citep{DialogueLLM:,Li_Liu_Qiao_Xu_2026}.

In RNN-based research, DS-LSTM   \citep{DS_LSTM} effectively handles context features at different times and frequencies and can process MFCC and Mel spectrograms separately. DialogueCRN \citep{DialogueCRN} introduces cognitive factors into emotion recognition tasks, providing a thorough understanding of cocontexts at the contextual and speaker level.

In Transformer-based research, EmoBERTa   \citep{EmoBERTa} extends RoBERTa  \citep{liu2019roberta} by adding speaker names before dialogues and inserting separators to predict the current speaker's emotion. For detailed content, please refer to Appendix \ref{Transformer-based Research}.

In GCN-based research, DialogueGCN  \citep{DialogueGCN} models dialogue utterances as graph nodes with context-based edges, improving emotion recognition through inter-speaker and self-dependencies. For detailed content, please refer to Appendix \ref{GCN-based Research}.

With the rapid development and widespread use of LLMs in emotion recognition, researchers are exploring how to leverage these models to enhance recognition effectiveness. InstructERC  \citep{lei2023instructerc} introduces an innovative approach that redefines the ERC task from a discriminative framework to a generative framework based on LLMs, guiding the generation of emotion classifications for specified categories. For detailed content, please refer to Appendix \ref{LLMs-based Research}.

\section{Methodology}
This section first introduces the research problem, then provides a detailed explanation of our proposed approach. For the feature extraction details, please refer to the appendix \ref{Emotion Label Extraction}.

\subsection{Problem Definition}

A conversation dataset can be defined as a sequence of utterances \(D\), where each utterance involves three aligned data sources corresponding to the acoustic (a), visual (v), and textual (t) modalities. This relationship can be represented by Equation \ref{equ:emo1}:
\begin{equation}
D_i = \{ u_{1}^{i}, u_{2}^{i}, u_{3}^{i}, \ldots, u_{N}^{i} \}
\label{equ:emo1}
\end{equation}

\(N \) represents the total number of utterances in the conversation, with \( i\) denoting one of the modalities: acoustic (a), visual (v), or textual (t). Our goal is to design a mapping \( F \) that transforms the input dialogue data \( D \) into a predefined set of emotion categories \( E \), where \( E \) includes emotions such as happiness, sadness, or anger.

\subsection{Transformer-GAT}
Figure \ref{fig:Transformer-GAT} illustrates the overall framework of our proposed dialogue emotion recognition system, Transformer-GAT, which consists of three main modules: the GAT module, the Transformer module, and the Emotion Classifier.  
\subsubsection{GAT Module}
\begin{figure*}[t]  
    \centering
    \includegraphics[width=0.9\textwidth]{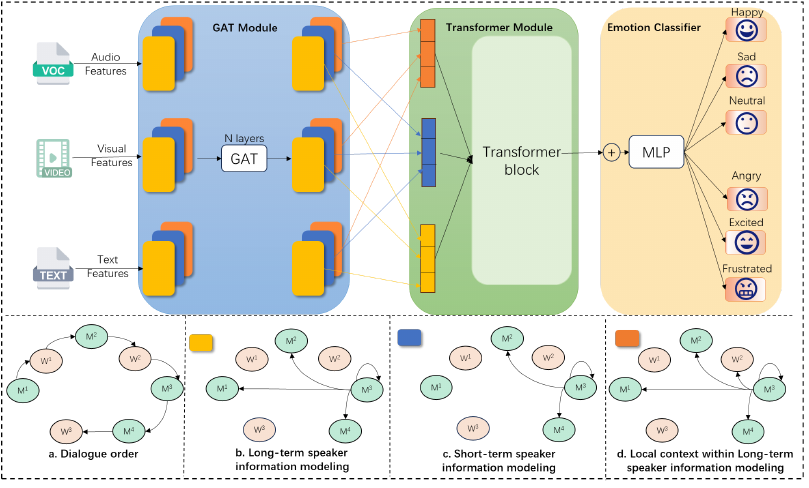}
    \caption{Transformer-based Graph Attention Network.}
    \label{fig:Transformer-GAT}
\end{figure*}
The module consists of three types of graph data with different structures, each designed to capture contextual information at various scales to comprehensively model the emotional features of dialogue. Specifically, we employ the LineConGraphs approach to model the relationships between other utterances and the target utterance. In this approach, edges are established between each relevant utterance and the target utterance, forming a linear graph. This way, each node is connected to the target utterance, allowing each graph to effectively capture contextual information in the dialogue under different topological structures.

GAT uses an attention mechanism to compute attention weights between nodes. The attention coefficient \( e_{ij} \) for nodes \( i \) and \( j \) is calculated as:
\begin{equation}\label{eq:4}
\small
e_{ij} = \text{LeakyReLU} \left( a^T \left[ W h_i \| W h_j \right] \right)
\end{equation}
where \( W \) is a learnable weight matrix, \( h_i \) and \( h_j \) are the feature vectors of nodes \( i \) and \( j \), respectively. The symbol \( \| \) denotes concatenation, and \( a \) is a learnable attention weight vector.

The attention coefficients \( e_{ij} \) are normalized using the softmax function, as shown in Equation \ref{eq:5}:
\begin{equation}\label{eq:5}
\alpha_{ij} = \frac{\exp(e_{ij})}{\sum_{k \in \mathcal{N}(i)} \exp(e_{ik})}
\end{equation}
where \( \mathcal{N}(i) \) is the set of neighboring nodes of node \( i \), and \( \alpha_{ij} \) is the normalized attention coefficient of node \( i \) to node \( j \).
Attention coefficients are used to weight the features of neighboring nodes, followed by feature aggregation, as shown in Equation \ref{eq:6}:
\begin{equation}\label{eq:6}
\small
h_i' = \text{ReLU} \left( \sum_{j \in \mathcal{N}(i)} \alpha_{ij} W h_j \right)
\end{equation}
where \(h_i'\) is the updated feature representation of node \(i\), \( \alpha_{ij} \) is the attention coefficient of node  \(j\) to node \(i\), \(W\) is the weight matrix for feature transformation, and \text{ReLU} is the activation function.

GAT typically uses multi-head attention to enhance the model's expressive power. Assuming there are \( k \) attention heads, the feature update for each head \( k \) is given by Equation \ref{eq:7}:
\begin{equation}\label{eq:7}
\small
h_i'^k = \text{ReLU} \left( \sum_{j \in \mathcal{N}(i)} \alpha_{ij}^k W^k h_j \right)
\end{equation}
The results from all heads are concatenated or averaged, as shown in Equation \ref{eq:8}:
\begin{equation}\label{eq:8}
h_i' = \frac{1}{K} \sum_{k=1}^K h_i'^k
\end{equation}
Each part of the network uses these differently structured graphs as input. This design not only enhances the model's ability to understand various contextual scales but also improves the accuracy and robustness of dialogue emotion recognition.
\subparagraph{Long-term speaker information modeling.}

In a dialogue, a person may continue to experience a particular emotion even as the context changes, a phenomenon known as emotional inertia \citep{liu2024emotionic}. Therefore, considering other utterances from the same speaker is crucial. To address this, we use an independent network to model each speaker's utterances in the dialogue. The multimodal features of each utterance are used as nodes in a graph. To fully account for other utterances by the same speaker, we connect each node in the graph to all other nodes belonging to the same speaker, constructing a fully connected graph. This constructed graph data is then input into the independent network to model specific long-term speaker information. Figure \ref{fig:Transformer-GAT} (a) shows the sequence of a dialogue, while Figure \ref{fig:Transformer-GAT} (b) illustrates how to use $M^3$ as the target utterance node and connect it to all other nodes of $M$’s utterances to construct edges. The method can be represented by Equation \ref{eq:9}:
\begin{equation}
\label{eq:9}
F_{\text{long}}^i = \text{ReLU}\left(\text{GAT}\left(U^i, \text{NodeList}\right)\right) 
\end{equation}
where \(U\) represents the target utterance, \(NodeList\) denotes the set of nodes corresponding to other relevant utterances, and \(i\) represents \(\{v, a, t\}\).
\subparagraph{Short-term speaker information modeling.}
The short-term same-speaker feature graph, as illustrated in Figure \ref{fig:Transformer-GAT} (c), captures the dynamic changes in emotions and context throughout the dialogue. By connecting each utterance, except for the first and last ones, to its preceding and following utterances, it enhances the model's understanding of contextual coherence, reduces information loss, and improves the ability to interpret the speaker's true intentions. Additionally, this approach supports dynamic emotional modeling, allowing the network to better reflect changes in the speaker's emotional state during the dialogue. The method can be represented by Equation \ref{eq:10}:
\begin{equation}\label{eq:10}
F_{\text{short}}^i = \text{ReLU}\left(\text{GAT}\left(U^i, \text{NodeList}\right)\right)
\end{equation}
\subparagraph{Local context within Long-term speaker information modeling.}
As shown in Figure \ref{fig:Transformer-GAT} (d), the graph extends from Figure \ref{fig:Transformer-GAT} (b) by adding the two nearest utterances surrounding the target utterance. Considering that local context emphasizes the most recent emotional information, while long-term context can indicate the overall emotional tone of the dialogue, this approach effectively tracks the sequence of utterances from the same speaker. By connecting adjacent utterances, it reduces information loss and ensures the context is maintained throughout the dialogue. These contexts complement each other, highlighting key emotions within the neighborhood while incorporating thematic emotional features. The method can be represented by Equation \ref{eq:11}:
\begin{equation}\label{eq:11}
F_{\text{local}}^i = \text{ReLU}\left(\text{GAT}\left(U^i, \text{NodeList}\right)\right)
\end{equation}
\subsubsection{Transformer Module}
After processing through the GAT block, we obtained graph data for long-term and short-term speaker information modeling, including local context features within long-term speaker information modeling. Transformer demonstrates significant advantages in multimodal fusion by leveraging self-attention mechanisms to capture long-range dependencies, dynamically model cross-modal interactions, handle variable-length inputs, and effectively enhance semantic representation accuracy. It comprehensively balances input features, reduces information loss, enriches feature representations, and significantly improves computational efficiency through its parallel processing architecture, making it highly effective for complex tasks. Based on the self-attention mechanism of the Transformer, we perform multimodal fusion for features within the same temporal or spatial scale, as described by the following formulas, where \(s\) represents feature types \{ \text{long}, \text{short}, \text{local} \}, and \(F\) represents multimodal feature combinations $\{ F_{\text{long}}^{all}, F_{\text{short}}^{all}, F_{\text{local}}^{all} \}$.

\begin{equation}\label{eq:12}
F_{\text{s}}^{all} = [F_{\text{s}}^a,F_{\text{s}}^v,F_{\text{s}}^t] 
\end{equation}
\begin{equation}\label{eq:13}
Q_{s},K_{s},V_{s} = WF+b 
\end{equation}
\begin{equation}\label{eq:14}
\text{weights}_{s} = \text{softmax} \left( \frac{Q_{s}K_{s}}{\sqrt{d_{k_s}}}  \right) 
\end{equation}
\begin{equation}\label{eq:15}
\text{output}_s = \text{weights}_s \cdot V_s
\end{equation}

\subsubsection{ Emotion Classifier}
After passing through the GAT block and the transformer block, the original multimodal features have been summarized into \(output_{long}\), \(output_{short}\) , and \(output_{local}\). Now, they are combined through the Emotion classifier, as shown in Equation \ref{eq:16}:
\begin{equation}\label{eq:16}
\small
output = output_{long}+output_{short}+output_{local}
\end{equation}
Subsequently, a multi-layer perceptron (MLP) is used to reduce the feature dimensions to match the number of emotion labels in the different datasets. Then, a LogSoftmax normalization function is applied to convert the model’s output into a probability distribution over various emotion categories, as shown in Equation 
\ref{eq:17}:
\begin{equation}\label{eq:17}
\small
 \widehat{Y} = argmax(LogSoftmax(W \dot output + b))
\end{equation}
After obtaining the model's output, the standard cross-entropy loss is used to compute the loss between the predicted labels and the true labels, as shown in Equation \ref{eq:18}:
\begin{equation}\label{eq:18}
\small
l =  -\frac{1}{N} \sum_{i=1}^{N} \sum_{j=1}^{C} Y_{ij} \log(\widehat{Y}_{ij})
\end{equation}
where \(N\) is the number of samples, \(C\)  is the number of emotion categories in the dataset, \(Y_{ij}\) is the true label of the \(i\)-th sample, \(\widehat{Y}_{ij}\) represents the probability that the model predicts the \(i\)-th sample belonging to class \(j\).
\section{Experimental Setup}
This section introduces experimental details, including partitioning of datasets, training specifics, and comparison models.
\subsection{Datasets}
IEMOCAP \cite{DBLP:journals/lre/BussoBLKMKCLN08} and MELD 
 \cite{Poria_Hazarika_Majumder_Naik_Cambria_Mihalcea_2019} are two key multimodal dialogue emotion datasets used for evaluating emotion analysis models. The IEMOCAP dataset includes multimodal data from movie dialogues (video, audio, text), focusing on dyadic conversations with emotion labels annotated by multiple experts. In contrast, the MELD dataset, derived from the TV show 'Friends' involves more complex dialogues with multiple speakers and a broader range of emotional categories. Both datasets provide diverse dialogue environments and emotion annotations, aiding in the comprehensive assessment of emotion analysis models. Details on dataset division are shown in Tables \ref{dataset-nums} and \ref{tab:data_statistics} , where No. Uttrs refers to the number of utterances and No. Dials refers to the number of dialogues. For emotion label classification, please refer to Appendix \ref{sec:emotion_label_classification}.
\begin{table}[htbp]
  \centering
  \small
  \begin{tabular}{cccc}
    \toprule
    \textbf{Dataset} & \textbf{Partition} & \textbf{No.Uttrs} & \textbf{No.Dials} \\
    \midrule
    \multirow{2}{*}{IEMOCAP} & train + val & 5810 & 120 \\
    & test & 1623 & 31 \\
    \multirow{2}{*}{MELD} & train + val & 11098 & 1152 \\
    & test & 2610 & 280 \\
    \bottomrule
  \end{tabular}
  \caption{Statistics of the two datasets}
  \label{dataset-nums}
\end{table}
\begin{table}[h]
	\small
	\centering
	\begin{tabular}{lccc}
		\toprule
		\textbf{Statistics} & \textbf{Train} &  \textbf{Dev}& \textbf{Test} \\
		\midrule
		Avg. Utt Length  & 15.15 / 8.03  & 17.00 / 7.99  & 20.10 / 8.28  \\
		Avg. Emo/Conv    & 4.86 / 3.30   & 5.10 / 3.35  & 4.97 / 3.24 \\
	  Avg. Utt/Conv    & 64.68 / 9.61  & 70.05 / 9.72  & 70.00 / 9.32 \\
		\bottomrule
	\end{tabular}
 \caption{ IEMOCAP and MELD datasets statistics. } \label{tab:data_statistics}
\end{table}
\subsection{Implementation Details}
In the experiments conducted on the IEMOCAP and MELD datasets, the learning rate for all networks was set to 1e-4, and the Adam optimizer was used. The hidden layer dimension was initialized at 256 and progressively halved. The feature dimensions for each modality are detailed in Table \ref{tab:Modal_Feature}. Both datasets were trained for 50 epochs on an NVIDIA GeForce RTX 4090 GPU. For each dataset, five different random seeds were employed, and the average results were reported as the final outcomes.
\begin{table}[h] 	
\small 	
\centering 	
\begin{tabular}{cccc} 		
\toprule 		
\textbf{Dataset} & \textbf{Text Dim} &  \textbf{Acoustic Dim}& \textbf{Visual Dim} \\ 		\midrule 		
IEMOCAP  & 768  & 100  & 256  \\ 		
  MELD    & 768   & 300  & 342 \\ 	  		\bottomrule 	
\end{tabular}  
\caption{ IEMOCAP and MELD Feature Dimensions. } 
\label{tab:Modal_Feature} 
\end{table}

\subsection{Evaluation Metrics}
Based on previous studies \cite{majumder2019dialoguernn}, we use the weighted average F1 score (Wa-F1) as the evaluation metric. This choice is driven by the issue of class imbalance, as observed in the IEMOCAP and MELD datasets, where the maximum difference in the number of emotion labels is 964 and 4815, respectively. In such cases, accuracy may not fully reflect the model’s actual performance, as the model might perform well in predicting majority classes while neglecting minority classes. Wa-F1 provides a more comprehensive assessment of overall classification performance by considering both precision and recall and weighting each class according to its importance. We also report detailed Wa-F1 scores for each class, As shown in the following equation:
\begin{equation}
\small
\label{eq:19}
\text{Accuracy} = \frac{\text{TP} + \text{TN}}{\text{TP} + \text{TN} + \text{FP} + \text{FN}}
\end{equation}
\begin{equation}
\small
\label{eq:20}
\text{F1} = \frac{2\text{TP}}{2\text{TP} + \text{FN} + \text{FP}}
\end{equation}
\begin{equation}
\small
\label{eq:21}
\text{Wa-F1} = \frac{\sum_{i} N_i \times F1_i}{\sum_{i} N_i}
\end{equation}
where \text{TP} stands for True Positives, \text{TN} stands for True Negatives, \text{FP} stands for False Positives, and \text{FN} stands for False Negatives. Furthermore, \( N_i \) represents the number of samples in class \( i \), and \( F1_i \) is the F1 score for class \( i \).

\section{Data analysis and visualization}
This section reports the experimental results and analyzes them, substantiating the performance of the model through visualization.
\subsection{Comparison with the State of the Art}
\begin{table*}[h]
  \centering
  \resizebox{\textwidth}{!}{%
  \begin{tabular}{lcccccccc}
    \toprule
    \textbf{Model} & \textbf{Happy} & \textbf{Sad} & \textbf{Neutral} & \textbf{Angry} & \textbf{Excited} & \textbf{Frustrated} & \textbf{Accuracy} & \textbf{Wa-F1} \\
    \midrule
    LineConGAT \cite{LineConGraphs}  & - & - & - & - & - & - & - & 64.58 \\
    MPLMM \cite{guo2024multimodal}  & - & - & - & - & - & - & 67.42 & 67.22 \\
    DualGATs \cite{DBLP:conf/acl/ZhangCC23} & - & - & - & - & - & - & - & 67.68 \\
    DAG-ERC \cite{DAG-ERC}& - & - & - & - & - & - & - & 68.03 \\
    LR-GCN \cite{DBLP:journals/tmm/RenHL0N22} & 55.50 & 79.10 & 63.80 & 69.00 & 74.00 & 68.90 & 68.50 & 68.30 \\
    EmoBERTa \cite{EmoBERTa} & - & - & - & - & - & - & - & 68.57 \\
    CBERL \cite{DBLP:journals/corr/abs-2312-06337} & 67.34 & 72.84 & 60.75 & \textbf{73.51} & 70.77 & 71.19 & 69.36 & 69.27 \\
    DER-GCN  \cite{DER-GCN} & 58.80 & 79.80 & 61.50 & 72.10 & 73.30 & 67.80 & 69.70 & 69.40 \\
    TelME \cite{yun2024telme} & 49.46 & 83.48 & 67.42 & 68.49 & 77.38 & 68.63 & - & 70.48 \\
    InstructERC \cite{lei2023instructerc}& - & - & - & - & - & - & - & 71.39 \\
    BERT-ERC \cite{Bert-erc} & - & - & - & - & - & - & - & 71.70 \\
    EmoCaps \cite{Emocaps} & \textbf{71.91} & 85.06 & 64.48 & 68.99 & \textbf{78.41} & 66.76 & - & 71.77 \\
    \textbf{Transformer-GAT(ours)} & 45.47 & \textbf{88.51$\uparrow$} & \textbf{78.51$\uparrow$} & 64.82 & 76.20 & \textbf{71.86$\uparrow$} & \textbf{73.04$\uparrow$} & \textbf{72.45$\uparrow$} \\
    \bottomrule
  \end{tabular}}
   \caption{Experimental results on the IEMOCAP dataset}
  \label{tab:emotion_performance_iemocap}
\end{table*}
Table \ref{tab:emotion_performance_iemocap} presents the experimental results in the IEMOCAP dataset, with bold fonts indicating the best performance. The results are sourced from the original papers, and "-" denotes missing results. Our model was tested with five random seeds, reporting the best performance. The results are sorted by the Wa-F1 score from low to high, and comparisons are made with current state-of-the-art models. Observations include:

\par 1) Our model surpasses the state-of-the-art, achieving a Wa-F1 score 0.68 higher than the 71.77 reported by EmoCaps.

\par 2) Among the six emotion categories, our model performs best in "Sad", "Neutral", and "Frustrated" with improvements of 3.45, 14.03, and 0.67, respectively. However, there are still gaps in the "Happy", "Angry", and "Excited" categories.
\begin{table*}[h]
  \centering
  \resizebox{\textwidth}{!}{%
  \begin{tabular}{lccccccccc}
    \toprule
    \textbf{Model} & \textbf{Neutral} & \textbf{Surprise} & \textbf{Fear} & \textbf{Sadness} & \textbf{Joy} & \textbf{Disgust} & \textbf{Anger} & \textbf{Accuracy} & \textbf{Wa-F1}\\
    \midrule
    DAG-ERC \cite{DAG-ERC} & - & - & - & - & - & - & - & - & 63.65 \\
    EmoCaps \cite{Emocaps} & 77.12 & 63.19 & 3.03 & 42.52 & 57.50 & 7.69 & 57.54 & - & 64.00 \\
    LR-GCN \cite{DBLP:journals/tmm/RenHL0N22}& 80.80 & 57.10 & 0.00 & 36.90 & 65.80 & 11.00 & 54.70 & - & 65.60 \\
    EmoBERTa \cite{EmoBERTa} & - & - & - & - & - & - & - & - & 65.61 \\
    DER-GCN \cite{DER-GCN}& 80.60 & 51.00 & 10.40 & 41.50 & 64.30 & 10.30 & 57.40 & 66.80 & 66.10 \\
    FacialMMT \cite{zheng-etal-2023-facial}& 80.13 & 59.63 & 19.18 & 41.99 & 64.88 & 18.18 & 56.00 & - & 66.58 \\
    CBERL \cite{DBLP:journals/corr/abs-2312-06337}& 82.03 & 57.91 & 22.23 & 41.36 & 65.67 & 24.65 & 55.31 & 67.78 & 66.89 \\
    DualGATs \cite{DBLP:conf/acl/ZhangCC23}& - & - & - & - & - & - & - & - & 66.90 \\
    BERT-ERC \cite{Bert-erc} & - & - & - & - & - & - & - & - & 67.11 \\
    TelME \cite{yun2024telme} & 80.22 & 60.33 & \textbf{26.97} & 43.35 & 65.67 & 26.42 & 56.70 & - & 67.37 \\
    InstructERC \cite{lei2023instructerc} & - & - & - & - & - & - & - & - & 69.15 \\
    LineConGAT \cite{LineConGraphs} & - & - & - & - & - & - & - & - & 76.50 \\
    \textbf{Transformer-GAT(ours)} & \textbf{87.53$\uparrow$} & \textbf{70.64$\uparrow$} & 25.29 & \textbf{59.41$\uparrow$} & \textbf{77.86$\uparrow$} & \textbf{33.98$\uparrow$} & 
     \textbf{72.80$\uparrow$} &\textbf{78.25$\uparrow$} &\textbf{77.37$\uparrow$}\\
    \bottomrule
  \end{tabular}}
  \caption{Experimental results on the MELD dataset}
  \label{tab:emotion_performance_meld}
\end{table*}

Table \ref{tab:emotion_performance_meld} presents the experimental results for the MELD dataset:
\par 1) Transformer-GAT shows excellent performance on the MELD dataset, with a Wa-F1 score of 0.87 higher than LineConGAT. Although EmoCaps and LineConGAT are the best performing models in their respective datasets, their performance is subpar in the other dataset. In contrast, our model achieves the highest Wa-F1 scores on both datasets and performs well across all emotion categories on the MELD dataset.

\par
\begin{figure}[h]
    \centering
    \includegraphics[width=0.4\textwidth, trim=0 30 0 30, clip]{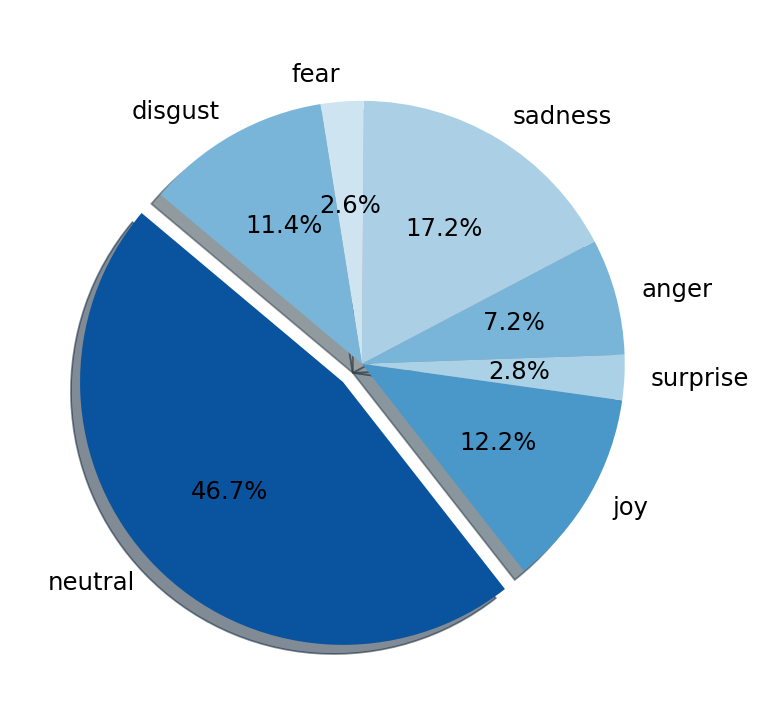}
    \caption{Comparison of the quantity of neutral emotion with other emotions.}  
    \label{fig:Emotion statistics}
\end{figure}

 2) In the MELD dataset, the distribution of emotion categories is significantly imbalanced, which presents substantial challenges compared to IEMOCAP, as shown in Figure \ref{fig:Emotion statistics}. This imbalance leads some models to focus more on the majority classes during training while neglecting the minority classes. Among the six models that provide detailed classification for all seven emotions (including Transformer-GAT), Transformer-GAT stands out with superior performance and greater adaptability. These experimental results and data analysis indicate that our model demonstrates strong adaptability to imbalanced datasets and achieves competitive performance, which is crucial for advancing the practical applications of emotion recognition technology.

\subsection{Transformer-GAT related experiments}
As shown in Table \ref{tab:different_numbers}, in experiments with stacked layers, the optimal number of layers is 2 for IEMOCAP and 1 for MELD. For IEMOCAP, the performance initially improves with additional layers, but then starts to decline. In contrast, for MELD, the performance initially decreases, with subsequent fluctuations observed as the depth increases. These results suggest that performance is sensitive to the number of layers and that the optimal layer count may vary between different datasets. This indicates the potential complexity of the modeling and highlights the need for further exploration to determine the ideal configuration for each data set.
\begin{table}[t]
    \centering
    \small
    \begin{tabular}{ccc}
        \toprule
        \textbf{GAT layers} & \textbf{IEMOCAP} & \textbf{MELD} \\
        \midrule
        1 & 70.74($\downarrow$1.71) & \textbf{77.37} \\
        2 & \textbf{72.45} & 73.46($\downarrow$3.91) \\
        3 & 71.70($\downarrow$0.75) & 68.87($\downarrow$8.50) \\
        4 & 71.65($\downarrow$0.80) & 70.28($\downarrow$7.09) \\
        \bottomrule
    \end{tabular}
\caption{Impact of Layer Number on Transformer-GAT Performance in ERC}
\label{tab:different_numbers}
\end{table}

\begin{figure}[h]
	\centering
	\includegraphics[width=0.5\textwidth]{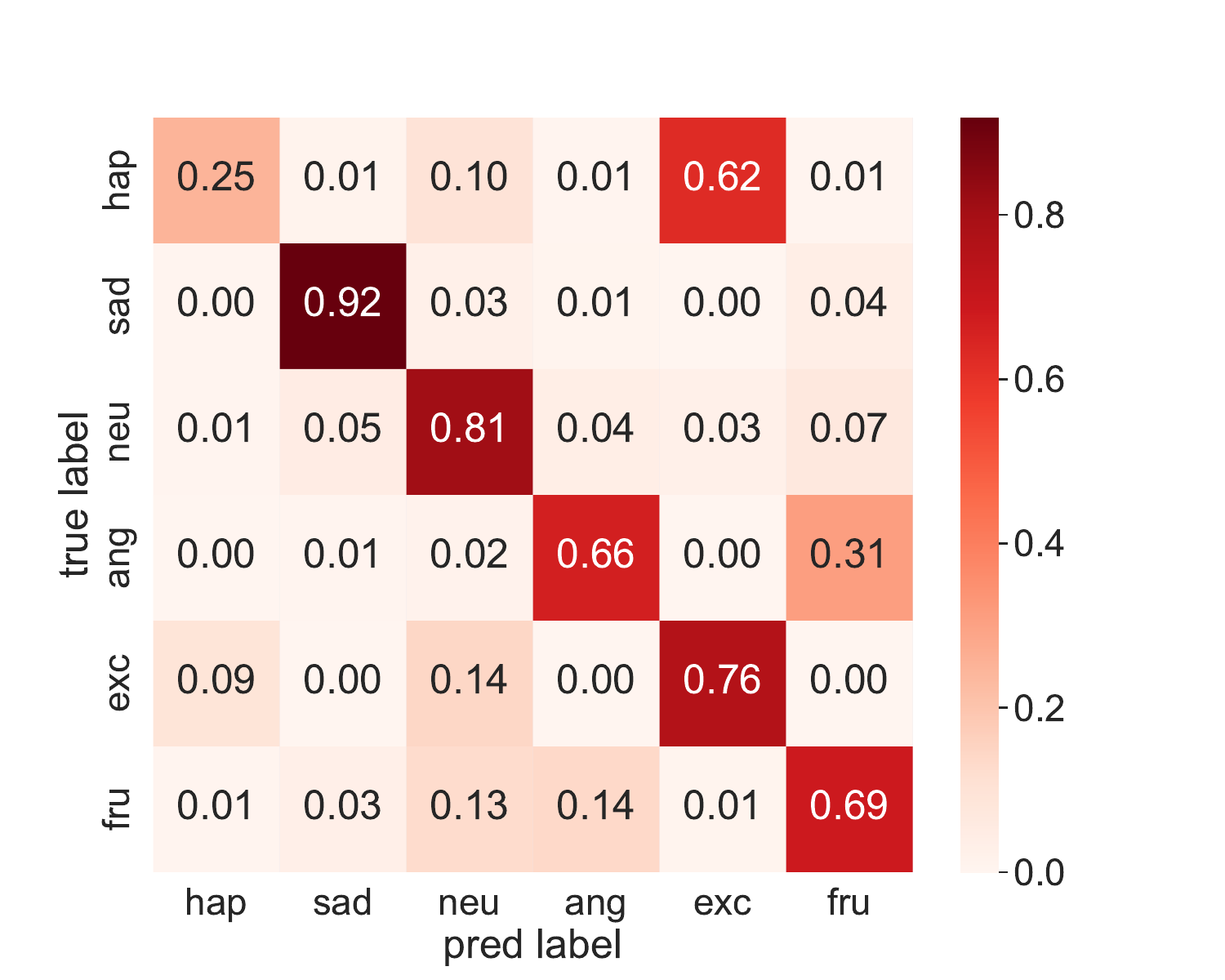}
	\caption{Confusion Matrix of IEMOCAP .}  
    \label{fig:iemocap}
\end{figure}

\begin{figure}[h]
	\centering
	\includegraphics[width=0.5\textwidth]{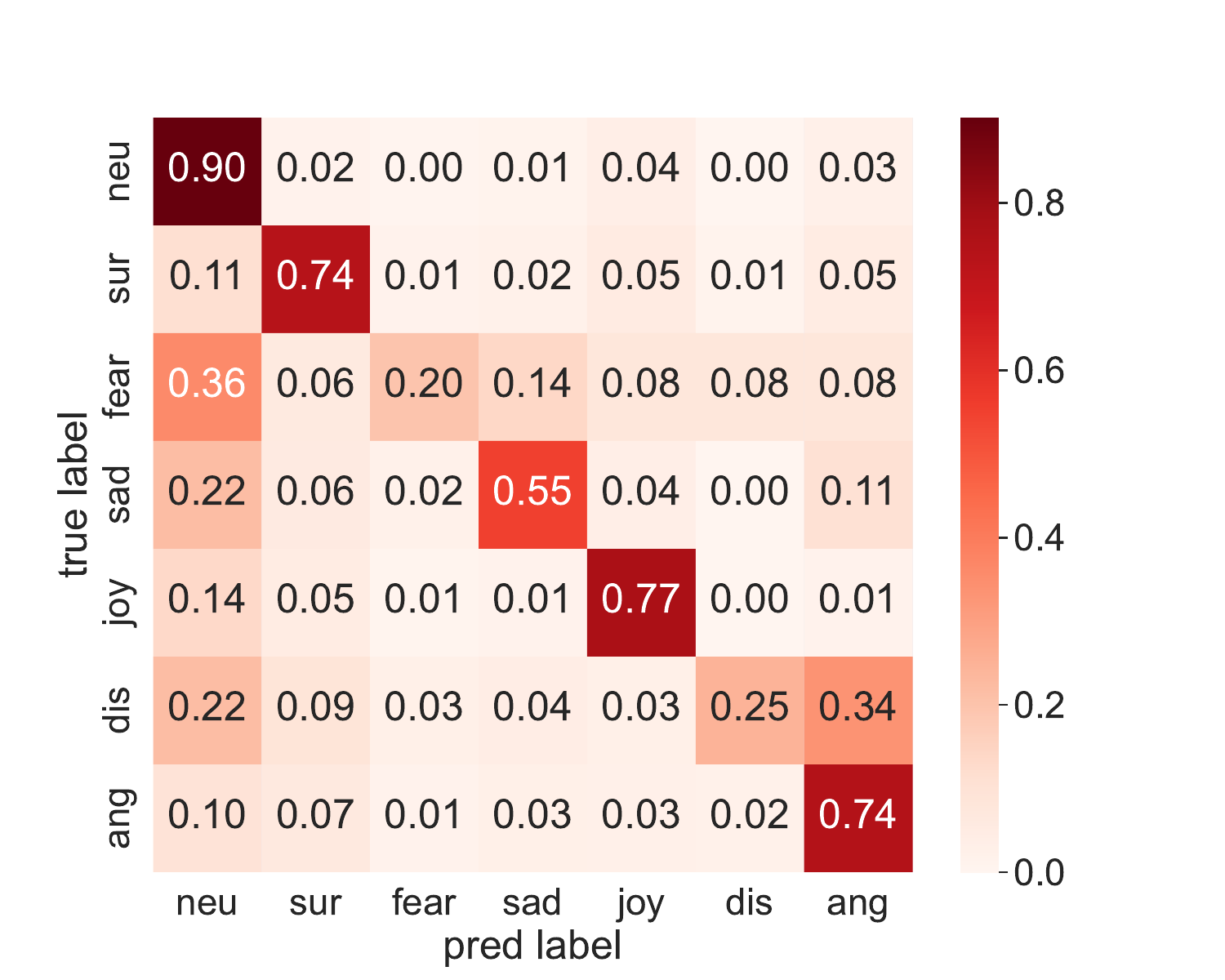}
	\caption{Confusion Matrix of MELD.}  
    \label{fig:meld}
\end{figure}

As shown in Figures \ref{fig:iemocap} and \ref{fig:meld}, the model performs well on most emotion labels, particularly with high accuracy for "Sadness" and "Anger" in the MELD dataset, and excellent performance for the "Neutral" emotion in the IEMOCAP dataset. However, the model encounters significant difficulties in recognizing minority emotion categories such as "Happy" and "Frustrated". Specifically, in the IEMOCAP dataset, the accuracy for "Happy" is as low as 0.25, and the performance in MELD is also suboptimal for this category. Additionally, there is substantial confusion between "Frustrated" and "Sadness" with the model frequently misclassifying these two emotions. The main reasons for these issues are the class imbalance in the datasets, where minority emotion categories (such as "Happy" and "Fear") have fewer samples, leading to insufficient learning for these classes. Furthermore, some emotions (like "Happy" and "Excited") share substantial similarities in expression, making it more challenging for the model to distinguish between them.

To provide a more intuitive demonstration of the model's performance across different emotion labels, the t-SNE plots allow us to observe the clustering of emotion categories in the MELD and IEMOCAP datasets, as well as potential overlaps and confusions between them. The detailed t-SNE plots can be found in the Appendix \ref{sec:tsne_appendix}. 
\subsection{Ablation Experiments}
We conducted two parts of ablation experiments. The first part evaluated the impact of multimodal data on Transformer-GAT's performance in ERC, while the second part examined the effect of three different graph structures on Transformer-GAT's ERC performance. The results were obtained by averaging the outcomes of five independent runs.

\par 1) As shown in Figure \ref{tab:multimodal}, among single-modal data, the video modality performed the worst, with weighted Wa-F1 scores of 14.33 and 31.27 for the two datasets, falling short of the best performance by 58.12 and 46.10. In contrast, the text modality achieved the best performance, with Wa-F1 scores of 71.57 and 75.69, only 0.88 and 1.68 below the best results. Among dual-modal combinations, the video and audio combination performed the worst, while the audio and text combination performed the best. However, the optimal results were obtained by combining all three modalities, with the IEMOCAP dataset scoring 72.45 and the MELD dataset reaching 77.37. This demonstrates that integrating all three modalities is crucial for emotion recognition tasks.
\begin{table}[h]
    \centering
    \small
    \begin{tabular}{ccc}
        \toprule
        \textbf{Modality} & \textbf{IEMOCAP} & \textbf{MELD} \\
        \midrule
        v & 14.33($\downarrow$58.12) & 31.27($\downarrow$46.10) \\
        a & 26.37($\downarrow$46.08) & 37.48($\downarrow$39.89) \\
        t & 71.57($\downarrow$0.88) & 75.69($\downarrow$1.68) \\
        va & 32.04($\downarrow$40.41) & 40.86($\downarrow$36.51) \\
        vt & 71.67($\downarrow$0.78) & 76.25($\downarrow$1.12) \\
        at & 71.73($\downarrow$0.72) & 76.40($\downarrow$0.97) \\
        vat & \textbf{72.45} & \textbf{77.37} \\
        \bottomrule
    \end{tabular}
    \caption{The effectiveness of Transformer-GAT for ERC under different modality combinations}
    \label{tab:multimodal}
\end{table}
\par 2) As shown in Table \ref{tab:graphs}, short-term speaker information modeling and Transformer are key factors affecting performance, with the Transformer having a more pronounced role in IEMOCAP. On the other hand, MELD shows lower sensitivity to long-term speaker information modeling, which may be attributed to its shorter dialogue length and structural characteristics. 
\begin{table}[h]
    \centering
    \small
    \begin{tabular}{ccc}
        \toprule
        \textbf{Different components} & \textbf{IEMOCAP} & \textbf{MELD} \\
        \midrule
        Transformer-GAT & \textbf{72.45} & \textbf{77.37} \\
        w/o Long-term speaker...& 71.47($\downarrow$0.98) & 76.96($\downarrow$0.41) \\
        w/o Short-term speaker... & 70.19($\downarrow$2.26) & 75.77($\downarrow$1.60) \\
       w/o Local context...  & 71.58($\downarrow$0.87) & 76.32($\downarrow$1.05) \\
       w/o transformer  & 70.39($\downarrow$2.06) & 76.99($\downarrow$0.38) \\
        \bottomrule
    \end{tabular}
    \caption{The impact of different graphs on Transformer-GAT's performance in ERC}
    \label{tab:graphs}
\end{table}

\section{Conclusion}
This study proposes a Transformer-based Graph Attention Network (Transformer-GAT) model for emotion recognition in dialogues, aiming to improve recognition accuracy. Experimental results show that the model achieves excellent performance on the IEMOCAP and MELD datasets, validating the effectiveness of the proposed approach. Despite relatively low Wa-F1 scores for some emotion labels, the overall model demonstrates strong generalization ability and high accuracy, effectively capturing emotion information and enhancing recognition performance, particularly in multi-modal feature fusion and emotion recognition tasks. 
\section*{Limitation}
The Transformer-GAT model performs well in most emotion classification tasks but faces challenges in distinguishing emotion categories with fewer samples. Future work will focus on optimizing the model structure, addressing class imbalance, and improving recognition of low-frequency emotions, while also validating the model in more complex dialogue scenarios.

\appendix
\newpage
\section{Dialogue Emotion Analysis}
\quad \textbf{Long-term speaker information modeling.} 
 From the long-term context, it is evident that the woman’s dissatisfaction has gradually accumulated during the discussion. Earlier in the conversation, she expressed anger when she said, "Oh no-no-no, give some specifics" and again when she said, "You fell asleep!" Both of these prior utterances show her frustration with the lack of attention and specific responses from the other party. As the conversation continues, her frustration builds up, culminating in the sentence, "They didn’t take any of my suggestions!" Here, her anger is directed not only at the man’s inattention but also at the fact that her suggestions were ignored. This progression illustrates the woman’s escalating emotional state, with the current sentence representing a peak of accumulated dissatisfaction and anger.

\textbf{Short-term speaker information modeling.}
The anger in the sentence "Woman: There was no kangaroo! (angry)" aligns with the emotions expressed in the preceding and following sentences, indicating the woman’s ongoing dissatisfaction with the lack of engagement from the other party. The previous sentence, "Woman: You fell asleep! (angry)" expresses her frustration with the man not paying attention to the movie, while the following sentence, "Woman: They didn’t take any of my suggestions! (angry)" further intensifies her anger, particularly about her suggestions being ignored. The middle sentence, "Woman: There was no kangaroo! (angry)" is a continuation of this emotional trajectory, reflecting her anger towards the man's inaccurate statement, showing the accumulation and intensification of her emotions.

\textbf{Local context within Long-term speaker information modeling.}

The anger in the sentence "Woman: You fell asleep! (angry)" originates from her strong dissatisfaction with the man's lack of attention to the film. The previous sentence, "Man: I was surprised to see a kangaroo in a World War I epic. (Surprise)" shows the man's surprise at the movie's content. The woman's anger is a response to the man's dishonesty and failure to pay attention to the movie. By denying the existence of the "kangaroo" scene, she expresses her frustration with his irresponsible behavior. This emotion is a direct reaction to the man's statement in the local context, indicating that her anger is primarily directed at his lack of engagement in the conversation.

\label{app:image_explanation}

\section{Supplementary Related Work}
\subsection{Transformer-based Research}
EmoCaps  \citep{Emocaps} proposes Emoformer, a Transformer-based structure that integrates emotion vectors from three modalities with sentence vector features. BERT-ERC  \citep{Bert-erc} enhances ERC performance with suggested text, fine-grained classification modules, and two-stage training, demonstrating excellent generalization capabilities. EACL \citep{EACL} is the first ERC model that addresses the emotion similarity problem by incorporating label semantic information, effectively guiding representation learning. CEPT \citep{gao2024cept} transforms the Emotion Recognition in Conversation task into a masked language modeling task through Prompt-Tuning, generating questions to effectively leverage pretrained language models (PLMs) for emotion recognition. Hicmae \citep{sun2024hicmae} provides a new solution for audio-visual emotion recognition tasks by combining self-supervised learning, hierarchical contrastive learning, and masked autoencoders.
\label{Transformer-based Research}
\subsection{GCN-based Research}
MMGCN \cite{MMGCN} treats multimodal features as nodes with connections within and between modalities. DAG-ERC \cite{DAG-ERC} incorporates speaker identity and positional information in Directed Acyclic Graph Neural Networks. LineConGraphs \cite{LineConGraphs} represents utterances through short-term contexts for emotion recognition. DER-GCN \cite{DER-GCN} examines the impact of event relationships on emotions by modeling contextual semantic information and dialogue relationships.
\label{GCN-based Research}

\subsection{LLMs-based Research}
OV-MERD \citep{openvocabulary} introduces an open lexicon for the first time, overcoming the previous limitation of predicting only a single emotion label. It now enables the generation of multiple emotion labels, better aligning with human emotional characteristics and emotional dynamics.
\label{LLMs-based Research}

\section{Multimodal Feature Extraction}
\label{Emotion Label Extraction}

 \quad\textbf{Acoustic Features:} we follow the methodology outlined by \citep{majumder2019dialoguernn} for acoustic feature extraction, utilizing the openSMILE toolkit \citep{eyben2010opensmile}. The extracted features are normalized and then subjected to dimensionality reduction via a fully connected layer.

\textbf{Visual Features:} we employ the DenseNet architecture \citep{huang2017densely}, pre-trained on the Facial Expression Recognition Plus (FER+) dataset \citep{barsoum2016training}, similar to the approach in \citep{MMGCN}.

\textbf{Text features:} we use the pre-trained EmoBERTa model, which has been trained on a large amount of sentiment-labeled data, making it more precise in capturing emotional nuances compared to general BERT models. We obtain the sentence representation by averaging the word embeddings from the final layer’s hidden states, as shown in Equations \ref{equ:emo2} and \ref{equ:emo3}:
\begin{equation}
\small
\text{output} = EmoBERTa(D_t)
\label{equ:emo2}
\end{equation}
\begin{equation}
\small
X_{text}=\text{MEAN}(\text{output}[\text{last\_hidden}])
\label{equ:emo3}
\end{equation}
where \(D_t\) represents the input dialogue, i.e., the text to be processed,
 and \(X_{\text{text}}\) denotes the text features extracted by the EmoBERTa model.

\section{Emotion Label Classification} \label{sec:emotion_label_classification}
\begin{figure}[h]
    \includegraphics[width=0.5\textwidth]{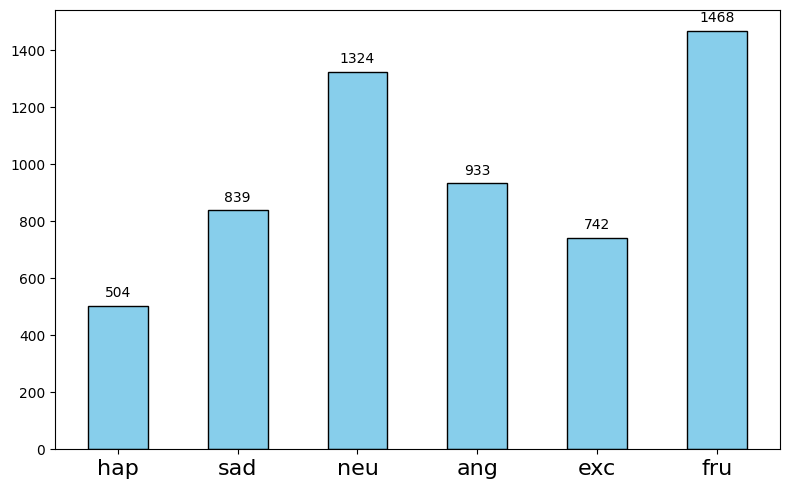}
    \caption{Emotion label distribution of IEMOCAP.}
    \label{fig:IEMOCAP_BAR}
\end{figure}

\begin{figure}[h]
    \includegraphics[width=0.5\textwidth]{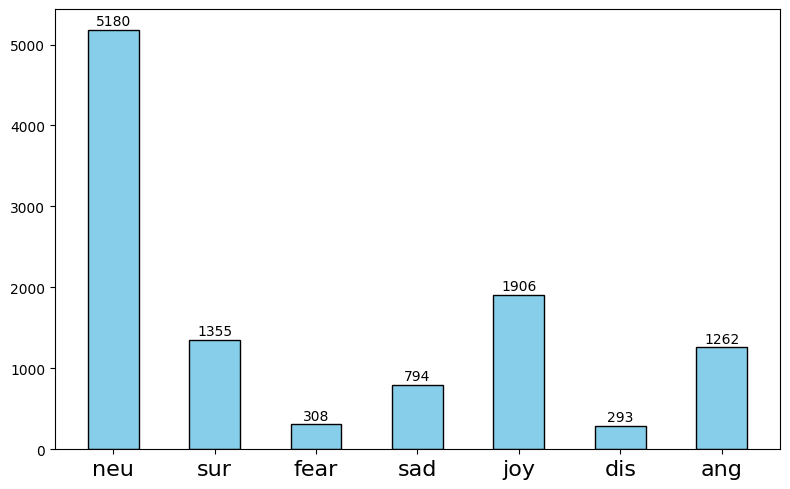}
    \caption{Emotion label distribution of MELD.}
    \label{fig:MELD_BAR}
\end{figure}
In both datasets, the minority emotion labels (such as "happy" and "fear") have a limited number of samples, resulting in a sparse distribution for these emotion labels.

\section{t-SNE Visualization of Emotion Labels}
\label{sec:tsne_appendix}
Although the Transformer-GAT model performs well in distinguishing most emotion labels, there is considerable confusion between certain categories, especially for minority labels such as "joy" and "fear". This confusion may be attributed to data imbalance and fine-grained differences between emotion labels. The limited training samples, the similarity between emotion categories, and the imbalanced sample distribution could all contribute to the model's poor performance on these labels.
\begin{figure}[htbp]
    \includegraphics[width=0.5\textwidth]{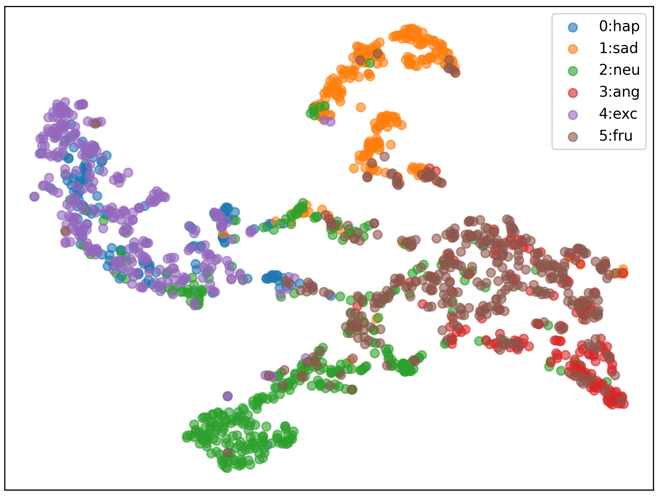}
    \caption{The t-SNE visualization of IEMOCAP.}
    \label{fig:IEMOCAP_TSNE}
\end{figure}
\begin{figure}[htbp]
    \includegraphics[width=0.5\textwidth]{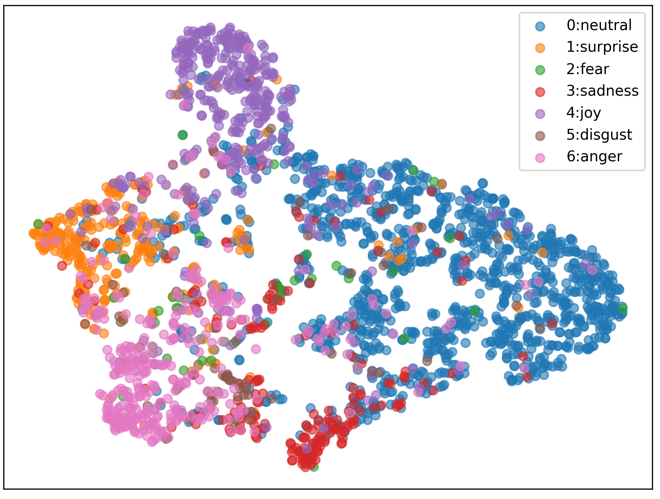}
    \caption{The t-SNE visualization of MELD.}
    \label{fig:MELD_TSNE}
\end{figure}
\end{document}